\documentclass{esannV2}
\usepackage[dvips]{graphicx}
\usepackage[utf8]{inputenc}
\usepackage{amssymb,amsmath,array}
\usepackage{natbib}
\usepackage{hyperref}
\usepackage{placeins}
\usepackage{lipsum}
\usepackage{url}
\usepackage{amsmath}
\usepackage[printonlyused,withpage,nolist]{acronym}
\usepackage[disable]{todonotes} 
\usepackage{subcaption}

\acresetall

\newcommand{\qq}[1]{\q{\q{#1}}}
\newcommand{\q}[1]{`#1'}

\newcommand{\warn}[1]{\PackageWarning{CustomWarning}{#1}}
\let\oldtodo\todo
\renewcommand{\todo}[1]{%
    \oldtodo[inline]{TODO: #1}
    \warn{TODO node placed: \qq{#1}}
}

\begin{document}
\title{Revisiting Neural Activation Coverage for Uncertainty Estimation}

\author{
Benedikt Franke\textsuperscript{\textdagger}$^1$, Nils~F\"{o}rster$^1$, Frank~K\"{o}ster$^1$, Asja Fischer$^2$, \\Markus~Lange$^1$ and Arne~Raulf$^1$\\
\textsuperscript{\textdagger} Corresponding author contact: \href{mailto:benedikt.franke@dlr.de}{benedikt.franke@dlr.de}.
%
%
\vspace{.3cm}\\
%
$^1$DLR Institute for AI Safety and Security, Ulm - Germany \\
$^2$Ruhr University Bochum, Bochum, Germany
}
%
\maketitle

\begin{abstract}
\Acf{NAC} is a recently-proposed technique for \acl{OoD} detection and generalization. 
We build upon this promising foundation and extend the method to work as an uncertainty estimation technique for already-trained artificial neural networks in the domain of regression. 
Our experiments confirm \ac{NAC} uncertainty scores to be more meaningful than other techniques, e.g.\ \acl{MC dropout}.
\end{abstract}

\acresetall

\section{Introduction}

For safety-critical applications, \ac{UE} of \acfp{ANN} is an important research area to counter-act overconfidence of wrong decisions and ensure trustworthiness of the predictions of the \ac{ANN}.
However, a lot of \ac{UE} techniques require intervention at training time, modifying the architecture of the \ac{ANN} to explicitly calculate uncertainty measures~\cite{gal2016dropout}.
This makes re-using pre-trained models challenging and may even impact the final performance of the model.
Wrapper-like \ac{UE} methods that can be applied to a trained \ac{ANN} without re-training or finetuning the neural network are only sparsely explored in the literature~\cite{oberdiek2018classification, ramalho2020density, ayhan2018test}.
To this end, \acf{NAC}, a technique to calculate an uncertainty measure based on the activation pattern of a trained \ac{CNN}, was recently introduced~\cite{liu2024neuron}.
In this work, we extend the method to work as an uncertainty estimator for regression, and verify the results by comparison with other \ac{UE} methods.
Our contributions can be summarized as follows:

\begin{enumerate}
    \item We extend the \ac{NAC} methodology to regression by proposing a new objective function.
    \item We confirm the competitiveness of \ac{NAC} against other uncertainty estimations techniques like \ac{MC dropout}~\cite{gal2016dropout} by direct comparison.
    \item We publicly release our code, including an optimized, easy-to-reuse, single-file adaptation\footnote{\url{https://github.com/DLR-KI/nac-uncertainty-regression}} of \ac{NAC} for PyTorch~\cite{paszke2019pytorch}.
\end{enumerate}



\section{Related Work}
\Acl{UE} in \ac{ANN}s has become a crucial area of research, especially for applications requiring reliable confidence measures.
One kind of uncertainty estimation focuses on calculating the uncertainty of \ac{ANN}s post-training, so that no modifications to the training protocol or network architecture are necessary.
These include approaches such as deep ensembles~\cite{lakshminarayanan2017simple}, which aggregate predictions from multiple independently trained networks, as well as methods based on test-time data augmentation~\cite{ayhan2018test}, which evaluate prediction variability by feeding augmented inputs through the fixed network.
Furthermore, calibration methods adjust output probabilities to better reflect uncertainty without altering the network weights~\cite{guo2017calibration}.

Another kind of uncertainty estimation methods operates during training, explicitly incorporating uncertainty modeling into the learning process~\cite{ganguly2021introduction, goan2020bayesian}. 
One representative of this concept is \Ac{MC dropout}~\cite{gal2016dropout}, where dropout is applied at both training and inference time to approximate Bayesian inference. 
By sampling multiple stochastic forward passes through the network at test time, \Ac{MC dropout} generates a distribution over predictions, providing an uncertainty estimate directly linked to the model’s learned parameters. 



\section{Extending Neural Activation Coverage for Uncertainty Estimation}
\subsection{Background}
We briefly restate the definition of \ac{NAC} as introduced in reference~\cite{liu2024neuron} with minor deviations of notation to clarify our adaptations.
Interested readers are referred to reference~\cite{liu2024neuron} for the more exhaustive introduction.
\Ac{NAC} calculates \textit{activation states} of selected neurons in an \ac{ANN} by backpropagating a \q{pseudo-loss}~$L$. This pseudo loss is given by the KL divergence between a uniform output vector~$u$ and the network output as softmax scores~$p$, thus we have
 $L(p) = D_{\text{KL}}(u || p)$.

Using the output~$z$ of one neuron of some intermediate layer and the pseudo-loss $L(p)$, the activation state~$\hat{z}$ is defined in Equation~\eqref{eq:activation-state}.
It is clamped between \texttt{0} and \texttt{1} by a sigmoid function~$\sigma$.

\begin{equation}
    \label{eq:activation-state}
    \hat{z} = \sigma(z \odot \frac{\partial L(p)}{\partial z})
\end{equation}

During calibration time, an approximate probability density function $\kappa^i_X$ is built up based on the computed $\hat{z}$-values of known \ac{ID} data~$X$ for each neuron~$i$ in the chosen layer(s).
This probability density function $\kappa^i_X$ is approximated as an histogram in actual implementation.
Using the parameter~$r>0$, defining the upper limit for each histogram bin, the \acs{NAC}-function is then defined as

\begin{equation}
    \Phi_X^i(\hat{z}_i; r) = \frac{1}{r} \text{min}(\kappa_X^i(\hat{z}_i), r)\,.
\end{equation}

Intuitively, $\Phi_X^i(\hat{z}_i; r)$ can be interpreted as a \qq{recognition score}, that increases the more often a neuron activation score has been observed during calibration phase.
For calculating an \ac{ANN}'s uncertainty, we consider it as a set of layers~$M$. 
For each~$m \in M$, the uncertainty scores of it's neurons are averaged.
To arrive at a final uncertainty score, the individual layer scores are summed. 
For this, we look up the respective activation scores~$\hat{z_i}$, according to Equation~\eqref{eq:activation-state}, and arrive at a set of activation states~$\hat{Z} = \{\hat{z_1}, \hat{z_2}, \ldots, \hat{z_n}\}$, with $n$ being the number of neurons in the network~\cite{liu2024neuron}.
We use the neuron-wise density functions~$\Phi_X^i$ for each neuron $i \in m$, to arrive at the final score

\begin{equation}
    S(M, X) = \sum_{m \in M} \frac{1}{|m|} \sum_{i \in m} \Phi_X^i(\hat{z}_i; r)\,.
\end{equation}

As we are interested in computing uncertainty scores, we take the inverse of the \acs{NAC} score~$U_{\text{NAC}} = S(M, X)^{-1}$, which is defined as long as ${M, X \ne \emptyset}$.

\subsection{\acs{NAC} for Regression}
For regression problems, computing the KL-Divergence~$L(p)$ is not possible, since the output of regression \acp{ANN} is one~(or more) scalar value(s), and not a probability vector.
However, since the uniform class probability vector $u = (1/C,\dots, 1/C)^\text{T}$ for $C$ classes is also the mean output given an \ac{i.i.d.} dataset, we propose a new pseudo-loss for regression problems.
Let $\bar{y}_X$ denote the mean output vector over some \ac{ID} calibration dataset $X$.
For a given vector $p$ of predictions of the network we utilize the \textit{Mahalanobis}~distance~\cite{mahalanobis2018generalized}~$d_{\text{m}}$ with  $L_X(p) = d_{\text{m}}(p, \bar{y}_X)$.

With this new definition for $L_X(p)$, \ac{NAC} can now be efficiently used for computing the uncertainty of regression problems. 
In Chapter \ref{sec:experiments}, we empirically evaluate this new definition.


\section{Experiments}
\label{sec:experiments}
To evaluate the suitability of \ac{NAC} towards assessing uncertainty specifically for detecting \ac{OoD} data, we train a 3-layer \ac{MLP} on ten regression datasets of the UCI~repository~\cite{kelly2023uci}.
The network has 128~neurons per hidden layer and uses SELU activations~\cite{klambauer2017self}.

\subsection{Out-of-Distribution Detection}
\label{sec:epistemic}
We generate synthetic artificial data by projecting each data point outside of the original distribution by adding noise drawn from a $\mathcal{N}(4\sigma, \frac{\sigma}{2})$ distribution where $\sigma$ is the standard deviation vector of the original dataset, i.e.,\ for a dataset originally distributed as $\mathcal{N}(\mu, \sigma)$.
Our \ac{OoD} data is distributed as $\mathcal{N}(\mu + 4\sigma, \frac{3\sigma}{2})$ and should therefore be sufficiently distinguishable from the original data.

\begin{figure}
    \centering
    \resizebox{0.7\textwidth}{!}{\input{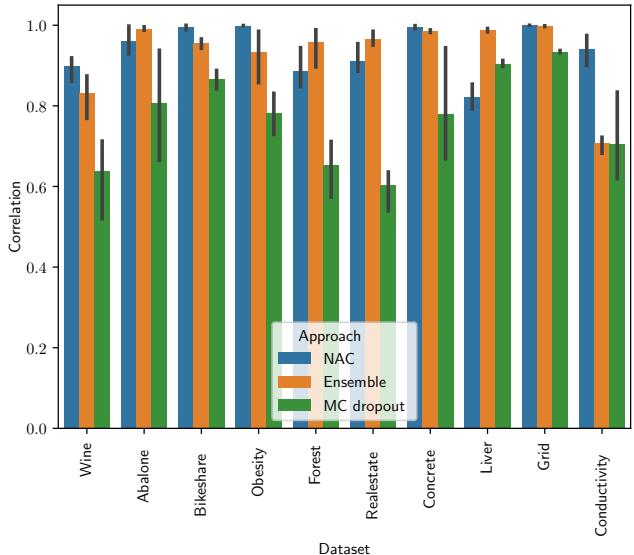}}
    \caption{
    \textbf{Comparing uncertainty approaches for \ac{OoD} detection.}
    Correlation of uncertainty value and \ac{OoD}-Label on ten~UCI regression datasets.
    \ac{NAC} adaptations~(ours) achieve the best correlation in six out of ten experiments.
    Error bars represent the 95\%~\acl{CI}.
    }
    \label{fig:epistemic}
\end{figure}

To verify our approach, we compare \ac{NAC} to established methods, such as ensembling, and \ac{MC dropout}.
For ensembling, we train ten networks simultaneously with bootstrapping data and take the standard deviation of the predictions as measure of uncertainty.
For \ac{MC dropout}, we insert a dropout layer after each linear layer and compute the uncertainty as the standard deviation of ten forward passes.
As \ac{NAC} introduces additional hyperparameters, we split off 10\% of the training data and generate an equal amount of \ac{OoD}-data for a quick hyperparameter sweep in each experiment after training the \ac{ANN}.
We compare the correlation between the uncertainty value and a binary marker for artificially created \ac{OoD}-data~(\texttt{0} for \ac{ID}, \texttt{1} for \ac{OoD}) across ten UCI-datasets for regression.
Our results are visualized in Figure~\ref{fig:epistemic}.

As depicted, \ac{NAC}'s uncertainty correlates best with the \ac{OoD} attribute in six out of ten experiments, which makes it the strongest uncertainty estimation technique of the three compared approaches.
In the four other cases it produces mostly comparative results to the other uncertainty estimation techniques.
These results show that \ac{NAC} is well-suited to compute uncertainty values for \ac{OoD} detection, even in the case of regression.

\subsection{Uncertainty Calibration}
\begin{figure}
    \centering
    \resizebox{0.7\textwidth}{!}{\input{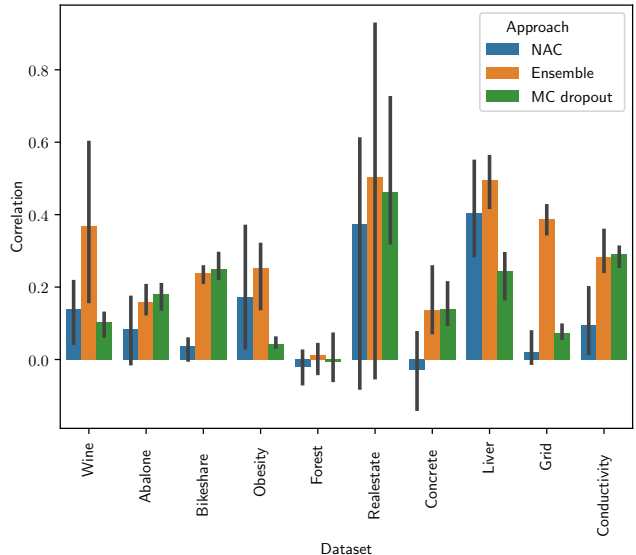}}
    \caption{
    \textbf{Comparing uncertainty approaches for \ac{ID} error detection.}
    Correlation of uncertainty value and per-sample \acs{MSE} on ten UCI regression datasets.
    NAC adaptations~(ours) achieve the smallest correlation in seven out of ten experiments.
    Error bars represent the 95\%~\acl{CI}.
    }
    \label{fig:aleatoric}
\end{figure}

As uncertainty can stem from a multitude of causes \cite{kendall2017uncertainties}, we want to find out if \ac{NAC} scores are influenced by errors on the \ac{ID} data. 
We repeat the previous comparison setup, however we do not generate any artificial data, but compute the correlation between the \acf{MSE} of the prediction on the test split and the uncertainty value.
Results are depicted in Figure~\ref{fig:aleatoric}.
In stark contrast to Figure~\ref{fig:epistemic}, we see that \ac{NAC} scores do not have a high predictive power for errors on \ac{ID} data, achieving the lowest correlation value of the three methods in seven out of ten cases.
This indicates a high calibration towards \ac{OoD}-driven uncertainty for \ac{NAC}-scores.

\section{Conclusion and Future Work}
In this work, we extended \ac{NAC} to regression and compared it with two other uncertainty techniques across ten~UCI datasets.
Our results show that \ac{NAC} is not only well-suited to compute uncertainty for \ac{OoD} detection.
In contrast, the approach tends to not be influenced by \ac{ID} prediction errors, enabling practitioners to efficiently compute uncertainty for \ac{OoD} detection.

We observed that \ac{NAC} is easier on computational resources when comparing the presented approaches, as both \ac{MC dropout} and ensembling need multiple forward passes, ensembling even requiring multiple trained models.
Like ensembling, it does not need any modification to trained \acp{ANN}, but can work on a single \ac{ANN}. 
This also applies to \ac{MC dropout} only if the trained network already contains dropout layers, which isn't the case in many modern applications of \acp{ANN}, e.g.\ modern deep \ac{CNN}s.
In future work, we aim to extend \ac{NAC} even further, e.g.\ towards Object Detection, which can be seen as a combination of classification and regression.

\begin{acronym}[MC Dropout]

 \acro{ANN}{artificial neural network}
 \acro{CI}{confidence interval}
 \acro{CNN}{convolutional neural network}
 \acro{ID}{in-distribution}
 \acro{i.i.d.}{independent and identically distributed}
 \acro{KL}{Kullback–Leibler}
 \acro{NAC}{neural activation coverage}
 \acro{MC dropout}{Monte-Carlo~Dropout}
 \acro{ODD}{operational~design~domain}
 \acro{OoD}{out-of-distribution}
 \acro{UE}{uncertainty estimation}
 \acro{MLP}{multi-layer Perceptron}
 \acro{MSE}{mean squared error}
 \acro{SELU}{scaled exponential linear units}

\end{acronym}


\begin{footnotesize}


\bibliographystyle{unsrt}
\bibliography{references_long}

\end{footnotesize}


\end{document}